\documentclass[runningheads]{llncs}
\usepackage[T1]{fontenc}
\usepackage{amsmath, amssymb}
\usepackage{graphicx,verbatim}
\usepackage{booktabs}
\usepackage{multirow}
\usepackage{color}
\usepackage{url}% 

\newcommand{\model}[0]{Dual-IFM}

\usepackage{adjustbox}
\usepackage{bm}

\usepackage{hyperref}
\hypersetup{
	colorlinks   = true, %Colours links instead of ugly boxes
	urlcolor     = blue, %Colour for external hyperlinks
	linkcolor    = blue, %Colour of internal links
	citecolor   = blue %Colour of citations
}
\makeatletter
\newcommand{\printfnsymbol}[1]{%
  \textsuperscript{\@fnsymbol{#1}}%
}
\begin{document}
\title{Towards Interpretable Foundation Models for Retinal Fundus Images}

% \begin{comment}  %% Removed for anonymized MICCAI 2025 submission
\author{
    Samuel Ofosu Mensah\inst{1,2}\thanks{These authors contributed equally.}\and %add \orcidID{} 
    Camila Roa\inst{1,2}\printfnsymbol{1,2} \and %add \orcidID{} 
    Kerol Djoumessi\inst{1} \and %add \orcidID{} 
    Philipp Berens\inst{1,2} %add \orcidID{} 
}
\authorrunning{S.O. Mensah et al.}
\institute{
    Hertie Institute for AI in Brain Health, University of T\"{u}bingen, Germany \and
    T\"{u}bingen AI Center, University of T\"{u}bingen, Germany\\
    \email{
        \{samuel.ofosu-mensah, maria-camila.roa-carvajal, philipp.berens \}@uni-tuebingen.de
    }
}

% \end{comment}

% \author{Anonymized Authors} 
% \authorrunning{Anonymized Authors et al.}
% \institute{Anonymized Affiliations \\
%     \email{email@anonymized.org}}

\maketitle              % typeset the header of the contribution
\begin{abstract}
Foundation models are used to extract transferable representations from large amounts of unlabeled data, typically via self-supervised learning (SSL). However, many of these models rely on architectures that offer limited interpretability, a critical issue in high-stakes domains such as medical imaging. We propose \model, a foundation model that is interpretable-by-design via a BagNet backbone whose small receptive fields generate class evidence maps that are faithful to the model's decision-making process. Additionally, \model{} incorporates a $2D$ projection layer during pretraining that enables direct visualization of the representation space, providing a dataset-level view of the learned structure including meaningful clinical clusters as well as potential spurious correlations. We trained \model{} on over 800,000 color fundus photographs from various sources to learn generalizable representations for different downstream tasks. Our model achieves performance comparable to RETFound, which has $16\times$ more parameters, while providing interpretable predictions on out-of-distribution data. These results suggest that large-scale SSL pretraining paired with inherent interpretability can lead to robust representations for retinal imaging. Code and pretrained models are available at \href{https://github.com/berenslab/interpretable_FM}{github.com/berenslab/interpretable\_FM}.

% Authors must provide keywords and are not allowed to remove this Keyword section.

\end{abstract}

\section{Introduction}
Foundation models are large-scale deep learning models trained on massive, diverse unlabeled datasets, typically via self-supervised learning (SSL), to learn general-purpose representations that can be adapted to downstream tasks~\cite{bommasani2021opportunities}. In ophthalmology, a widely used foundation model called RETFound~\cite{zhou2023foundation} has inspired many studies across ophthalmic imaging tasks~\cite{du2024ret,shi2025multimodal,sun2025data}. While these models typically achieve strong performance, they are not inherently interpretable, offering no built-in explanation for each prediction~\cite{grote2023allure,kazmierczak2025explainability}. Thus, several studies on foundation models rely on post-hoc attribution methods to explain the model's decisions~\cite{sun2025data,shi2025multimodal,zhou2023foundation}. However, these local explanations are often unfaithful, since they do not align with the model's decision-making process~\cite{adebayo2018sanity,rudin2019stop}. This lack of faithfulness limits the adoption of foundation models, especially in high-stakes domains such as medical imaging.

Understanding the model's decision-making process is important, so as the semantic structure of the representations it learns from the data~\cite{bengio2013representation}. Foundation models yield high-dimensional representations of input data, embedding semantic relationships in the geometric properties of the latent space. Although attribution methods aim to provide local explanations for individual predictions, they do not reveal whether the underlying representation space accurately captures semantic relationships, that is, whether semantically similar concepts such as disease grades are proximally embedded and whether the geometric structure reflects meaningful patterns rather than data artifacts or spurious correlations~\cite{achtibat2023attribution}. Visualizing representation spaces offers a complementary dataset-level view of the learned global structure, typically done via neighbor embedding methods such as \textit{t}-SNE~\cite{kobak2019art,sun2025data}. However, \textit{t}-SNE is non-parametric: it learns no explicit high-to-low-dimensional mapping and must be re-fitted whenever new samples are added, which can be costly for large datasets.

Motivated by these considerations, we introduce \model, an interpretable foundation model for color fundus photography (CFP) that combines local explanations with a direct visualization of the representation space. We trained our model using the BagNet~\cite{brendel2019approximating} architecture with the \textit{t}-SimCNE algorithm~\cite{bohm2022unsupervised}. Unlike post-hoc attribution methods, BagNet is interpretable-by-design: by restricting the receptive field to small patches, it generates class evidence maps that indicate which regions of the image contributed to a specific prediction. \textit{t}-SimCNE trains a parametric projection layer that maps images into a $2D$ approximation of the representation space, enabling \model{} to embed new images directly without re-fitting non-parametric methods such as  \textit{t}-SNE. Together, these components constitute a framework for local interpretability and dataset-level representation inspection of the learned structure, supporting more transparent use of foundation models in high-stakes domains like ophthalmology.

\section{Methods}
We trained the inherently-interpretable foundation model \model{} on a large dataset of CFP. The model has a BagNet architecture as the backbone~\cite{brendel2019approximating} and is trained to learn generalizable representations using the self-supervised contrastive \textit{t}-SimCNE algorithm~\cite{bohm2022unsupervised}.

\begin{figure}[t]
    \centering
    \includegraphics[width=\textwidth]{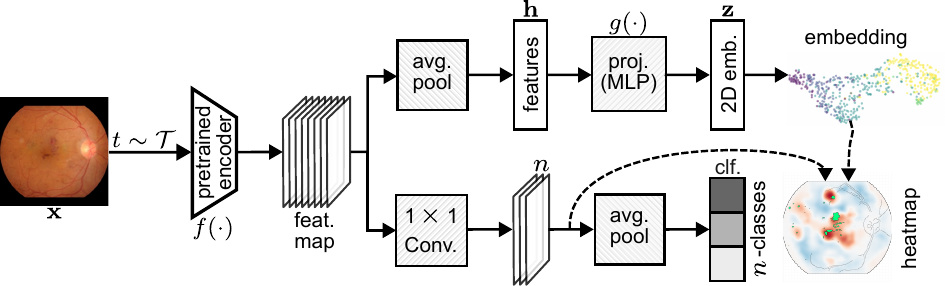}
    \caption{
    \textbf{Architecture of \model}. The BagNet encoder is pretrained on a large CFP dataset with \textit{t}-SimCNE, learning generalizable representations and a $2D$ projection of the embedding space. A classification branch can be added for fine-tuning on downstream tasks while preserving the $2D$ projection.
    }
    \label{fig:overview}
\end{figure}

\subsection{Architecture}
Fig.~\ref{fig:overview} shows the architecture of \model, where the upper branch is learned during pretraining. We decompose it into an encoder $f:\mathcal{X} \to \mathcal{H} \subset \mathbb{R}^D$, with $D$ denoting the encoder feature dimension, and a projection layer $g : \mathcal{H} \to \mathcal{Z} \subset \mathbb{R}^2$. The encoder produces a spatial feature map which is aggregated by global average pooling to obtain a $D$-dimensional feature vector $\mathbf{h}$, and the projection head maps $\mathbf{h}$ to a $2D$ embedding $\mathbf{z} = g \circ f(\mathbf{x})$, at which the contrastive loss is computed. Following SimCLR~\cite{chen2020simple}, the high-dimensional representations $\mathbf{h}$ of the encoder are used for downstream tasks, while the projection head is retained to embed images into a $2D$ visualization of the representation space.

After pretraining, we introduce a classification head to form the lower branch (Fig.~\ref{fig:overview}), operating on the encoder's feature map in parallel with the projection layer. Following~\cite{donteu2023sparse}, we attach a $1 \times 1$ convolution to the final feature map to produce $n$ class-wise activation maps that directly highlight the regions of the image that contributed to the model's predictions. Global average pooling is then applied over each activation map to yield a corresponding class score, making the prediction an explicit spatial aggregation of local evidence. This classification head can be used for fine-tuning on downstream tasks.

\subsection{Pretraining}
We trained \model{} using the \textit{t}-SimCNE algorithm~\cite{bohm2022unsupervised}, a variant of SimCLR~\cite{chen2020simple}, that adapts the contrastive objective to obtain a $2D$ visualization of the embedding space through a three-stage training procedure. It replaces SimCLR's $128D$ projection, on which the contrastive loss is computed, wit a $2D$ projection, directly embedding the input images in a $2D$ space. Additionally, \textit{t}-SimCNE replaces the cosine similarity, used by SimCLR's NT-Xent loss with Euclidean similarity to improve visualization quality. In stage one, the full model is trained computing the loss at the $128D$ output of the projector. In stage two, the last projector layer is replaced with a $2D$ output. This layer is then aligned with the encoder's representations, while the other layers remain frozen. In stage three, the full model is unfrozen and trained end-to-end at a lower learning rate.

The authors of \textit{t}-SimCNE found that using Euclidean similarity throughout all stages yields the best visualizations, but also reported that cosine similarity during stage one improves linear probe performance at a slight cost to visualization quality. Evaluating both, we observed the opposite effect on visualization quality (see Section~\ref{results-global}) and therefore adopted cosine similarity in stage one.

\subsection{Fine-tuning}
During fine-tuning, the encoder and classification head are trained end-to-end while the projector is kept frozen. As the encoder representations are updated, the $2D$ projector's mapping may become suboptimal, with the degree of degradation depending on how much the representations drift from the pretraining ones. When the representations remain close to pretraining, the projector remains a reliable approximation; otherwise, it can be realigned to the fine-tuned representations in a computationally inexpensive way, since only the projector is retrained. This alignment can be performed during or after fine-tuning.

The faithfulness of the activation maps produced by the lower branch stems from the BagNet backbone~\cite{brendel2019approximating}, which restricts the receptive field of each spatial location to a small patch of the input image. This means each entry in the feature map depends only on a local region of the image, ensuring that the class evidence maps directly reflect the local image content that contributed to the prediction. Unlike post-hoc attribution methods, which approximate explanations after inference, this interpretability is built into the architecture itself. The evidence maps are not an explanation layered on top of the decision process, but an intrinsic part of it, eliminating the need for any post-hoc attribution methods.

\subsection{Experimental details}
We pretrained \model{} on three CFP datasets that were preprocessed and filtered for quality~\cite{Gervelmeyer2025-fit}: EyePACS (567,384 images from 44,063 participants)~\cite{cuadros2009eyepacs}, AREDS (110,690 images from 4,432 participants)~\cite{age1999age}, and UKBiobank (132,010 images from 72,711 participants)~\cite{warwick2023uk}, yielding 802,360 images. We evaluated the model on out-of-distribution datasets: APTOS~\cite{goha2024aptos} (3,662 images), IDRiD~\cite{porwal2018indian} (516 images), DeepDRiD~\cite{liu2022deepdrid} (2,000 images) and Messidor-2~\cite{decenciere2014feedback} (1,718 images) for DR grading, Glaucoma fundus~\cite{ahn2018deep} (1,544) and PAPILA~\cite{kovalyk2022papila} (488) for glaucoma grading, and FIVES~\cite{jin2022fives} for multi-disease classification. We used BagNet-33 (variant with a $33\times33$-pixel receptive field) as \model{}'s backbone and compared it with a BagNet-33 initialized with ImageNet weights, a BagNet-33 trained with SimCLR on our pretraining dataset, and RETFound (ViT-Large).

For pretraining, we used the LARS optimizer~\cite{you2017large} (batch size 1024, weight decay $10^{-6}$, base learning rate $0.075 \times 32$). The learning rate was reduced by a factor of $1000$ in the last stage, following a cosine annealing schedule with a warm-up of 10 epochs for stages one and three, and kept constant for stage two. Both models were trained with mixed precision. While SimCLR is compatible with float16, we found that \textit{t}-SimCNE requires bfloat16 on large datasets, as the unconstrained Euclidean distances in the loss can otherwise lead to overflow.

We trained a SimCLR baseline for $1000$ epochs and used these weights to initialize the \textit{t}-SimCNE variant with cosine similarity in stage one, which was then trained for $25$ and $200$ epochs in stages two and three, respectively. \textit{t}-SimCNE with Euclidean similarity was trained from scratch for $775$, $25$, and $200$ epochs. As we show in Section~\ref{results-global}, the cosine variant yielded better visualizations for our CFP dataset and was therefore adopted as \model. Pretraining took up to 6 days using 8 NVIDIA H100 GPUs (see Table~\ref{tab:compute-cost}).

Linear probing was performed by fitting a logistic regression classifier with elastic net penalty and class-balanced weights on the model's representations for each dataset, and evaluating performance on a held-out test set. For fine-tuning, we trained the linear classification head for 10 epochs before unfreezing the full model, continuing for a maximum of 50 epochs with early stopping. We used class-weighted cross entropy and five-fold cross-validation and report average performance across folds. For all datasets, we ensured that splits are performed at participant-level with stratification. Hyperparameters were selected via a small search for each dataset and full configurations are provided in our GitHub repository. We conducted the fine-tuning on one NVIDIA V100 GPU.

\begin{table}[b!]
    \centering
    \caption{Model complexity and computational cost. Pretraining figures for RETFound are as reported in~\cite{zhou2023foundation} (*described as "developing time" rather than training wall-clock); all other figures were measured in this work.}
    \label{tab:compute-cost}
    \begin{tabular}{lcc}
        \toprule
        & \model & RETFound \\
        \midrule
        Architecture & BagNet-33 & ViT-Large \\
        \# Parameters & 18.3M & 303.3M \\
        \addlinespace
        \multicolumn{3}{l}{\textbf{Pretraining}} \\
        Data (\# images) & 802,360 & 1.6M (900k private) \\
        Hardware & 8$\times$H100 (80GB) & 8$\times$A100 (40GB) \\
        Epochs & 1225 (1000+25+200) & 800 \\
        Time (effective batch size) & 6d 8h (1024) & 2 weeks* (1{,}792) \\
        \addlinespace
        \textbf{Inference} (batch size 16, V100, fp32) & & \\
        Time per batch & 81 ms & 140 ms \\
        Peak memory & 1.12 GB & 1.31 GB \\
        \bottomrule
    \end{tabular}
\end{table}

\begin{table}[b!]
    \centering
    \caption{Linear-probe performance across various CFP tasks with model properties.}
    \label{tab:auroc-linear}
    \resizebox{\linewidth}{!}{%
        \begin{tabular}{lccccc}
            \toprule
            & ImageNet & SimCLR & RETFound & \textit{t}-SimCNE & \model \\
            \midrule
            Image-level interpretability & $+$ & $+$ & $-$ & $+$ & $+$ \\
            Dataset-level structure & $-$ & $-$ & $-$ & $+$ & $+$ \\
            In-Domain training & $-$ & $+$ & $+$ & $+$ & $+$ \\
            \midrule
            EyePACS & $0.727$ & $0.647$ & $\underline{0.794}$ & $0.648$ & $\textbf{0.849}$ \\
            AREDS & $0.877$ & $\underline{0.925}$ & $0.871$ & $\textbf{0.936}$ & $0.919$ \\
            APTOS & $\textbf{0.931}$ & $\underline{0.929}$ & $0.917$ & $\underline{0.929}$ & $0.925$ \\
            DeepDRiD & $0.854$ & $0.857$ & $\textbf{0.887}$ & $\underline{0.861}$ & $0.829$ \\
            IDRiD & $0.702$ & $\underline{0.762}$ & $0.668$ & $\textbf{0.785}$ & $0.758$ \\
            Messidor & $0.801$ & $\textbf{0.868}$ & $0.797$ & $\textbf{0.868}$ & $\underline{0.820}$ \\
            Glaucoma & $0.908$ & $\textbf{0.921}$ & $0.913$ & $0.912$ & $\underline{0.915}$ \\
            PAPILA & $0.655$ & $\underline{0.740}$ & $0.577$ & $0.682$ & $\textbf{0.757}$ \\
            FIVES & $0.895$ & $\underline{0.912}$ & $0.837$ & $\textbf{0.917}$ & $0.881$ \\
            \bottomrule
        \end{tabular}
    }
\end{table}

\section{Results}~\label{results}
\subsection{Performance on eye disease classification}
On held-out test sets of the pretraining datasets, \model{} performed comparably to RETFound on in-distribution DR detection and AMD classification. On out-of-distribution datasets, \model{} matched or outperformed a task-trained BagNet-33 across most tasks (Tables~\ref{tab:auroc-linear} and \ref{tab:auroc-finetune}), indicating that self-supervised domain-specific pretraining improves performance on downstream tasks. Additionally, we found that \model{} performed similarly to SimCLR and RETFound, suggesting that the added interpretability and $2D$ projection do not impair representation quality. Linear probing results (Table~\ref{tab:auroc-linear}) showed improved representations for 4/7 and 6/7 datasets compared to the task-trained BagNet baseline (ImageNet initialized) and RETFound, respectively. Fine-tuning further improved performance for most datasets and models (Table~\ref{tab:auroc-finetune}), with the exception of PAPILA and FIVES, where performance occasionally decreased, likely due to overfitting given the limited hyperparameter search. Despite this limitation, \model{} performs on par with or above RETFound, suggesting that it learns general representations from retinal images.

\begin{table}[b!]
    \centering
    \caption{Fine-tuning performance across 5-folds on various CFP tasks.}
    \label{tab:auroc-finetune}
    \resizebox{\linewidth}{!}{%
        \begin{tabular}{lccccc}
            \toprule
            & ImageNet & SimCLR & RETFound & \textit{t}-SimCNE & \model \\
            \midrule
            EyePACS & $0.758 \pm 0.073$ & $\underline{0.804 \pm 0.038}$ & $0.767 \pm 0.040$ & $0.759 \pm 0.037$ & $\bm{0.821 \pm 0.036}$ \\
            AREDS & $0.908 \pm 0.020$ & $\underline{0.924 \pm 0.008}$ & $0.912 \pm 0.008$ & $\bm{0.936 \pm 0.007}$ & $0.923 \pm 0.009$ \\
            APTOS & $0.938 \pm 0.003$ & $\bm{0.941 \pm 0.009}$ & $\underline{0.940 \pm 0.004}$ & $0.935 \pm 0.010$ & $\underline{0.940 \pm 0.003}$ \\
            DeepDRiD & $0.889 \pm 0.008$ & $\underline{0.901 \pm 0.014}$ & $\bm{0.903 \pm 0.007}$ & $0.883 \pm 0.006$ & $0.891 \pm 0.012$ \\
            IDRiD & $\underline{0.807 \pm 0.033}$ & $\bm{0.814 \pm 0.029}$ & $0.764 \pm 0.032$ & $0.796 \pm 0.018$ & $0.766 \pm 0.030$ \\
            Messidor & $\underline{0.865 \pm 0.020}$ & $\bm{0.874 \pm 0.006}$ & $0.845 \pm 0.008$ & $0.860 \pm 0.012$ & $0.851 \pm 0.038$ \\
            Glaucoma & $0.912 \pm 0.013$ & $0.930 \pm 0.007$ & $\bm{0.947 \pm 0.007}$ & $\underline{0.933 \pm 0.009}$ & $0.925 \pm 0.012$ \\
            PAPILA & $0.648 \pm 0.056$ & $0.664 \pm 0.061$ & $\bm{0.689 \pm 0.058}$ & $0.676 \pm 0.051$ & $\underline{0.680 \pm 0.027}$ \\
            FIVES & $0.881 \pm 0.021$ & $0.891 \pm 0.015$ & $\bm{0.936 \pm 0.012}$ & $\underline{0.910 \pm 0.024}$ & $0.898 \pm 0.030$ \\
            \bottomrule
        \end{tabular}
    }
\end{table}

\subsection{Global structure inspection through $2D$ projection}~\label{results-global}
\begin{figure}[t]
\centering
% \begin{minipage}{0.95\textwidth}
\centering
\includegraphics[width=0.8\textwidth]{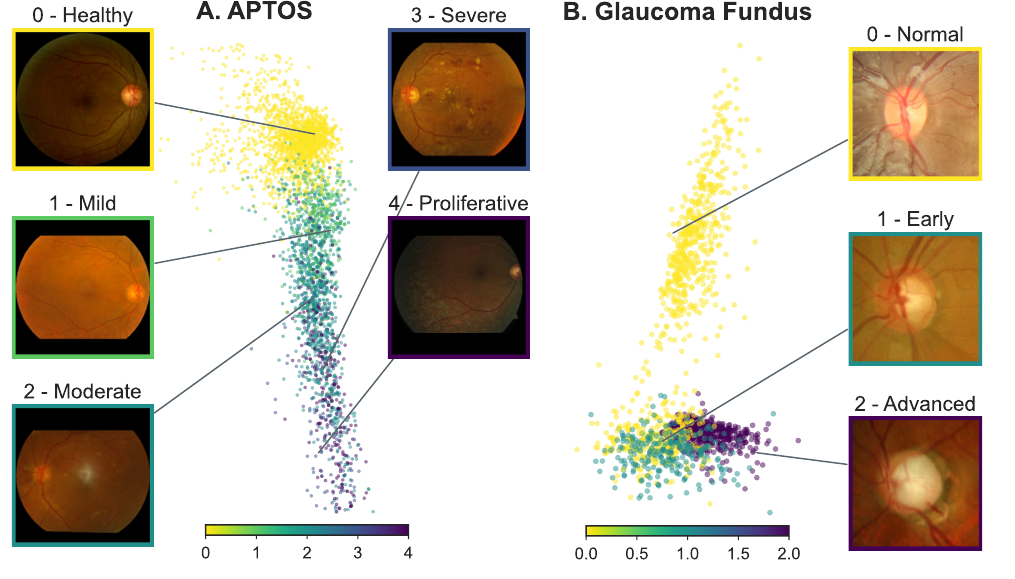}
\vspace{0.5em}
\caption{\textbf{Representation space visualization}. $2D$ embeddings of (A) APTOS (\textit{k}-NN AUROC: $0.881$) and (B) Glaucoma Fundus (\textit{k}-NN AUROC: $0.883$), revealing disease progression structure and potential borderline cases.}
\label{fig:emb}
\end{figure}

Using the learned $2D$ projection, \model{} enables dataset-level inspection of the representation space by embedding entire datasets into a low-dimensional approximation. Fig.~\ref{fig:emb} shows two datasets as an example, illustrating how the representation space captures disease severity structure. For the APTOS dataset, samples arranged along a curved manifold following DR severity: images of lower grades clustered in one region, while progressively higher grades occupied increasingly distinct regions~(Fig.~\ref{fig:emb}A, yellow to blue). The embedding for Glaucoma Fundus exhibited a similar structure~(Fig.~\ref{fig:emb}B). For both datasets, we observed overlap among disease grades, highlighting potential borderline cases.

To quantify the quality of the $2D$ embeddings, we measured the \textit{k}-NN AUROC across all datasets, comparing \model{} to \textit{t}-SNE and $2D$ PCA applied post-hoc to the high-dimensional representations of the baselines (Table~\ref{tab:auroc-finetune}). Contrary to the findings of the \textit{t}-SimCNE, using cosine similarity in stage one achieved better $2D$ embedding quality than Euclidean similarity throughout ($0.821$ vs $0.745$), leading us to adopt this variant as \model. The $2D$ visualization can be further improved by aligning the projector to the fine-tuned encoder representations. However, this alignment is not always necessary, as illustrated by the meaningful structure already visible in Fig.~\ref{fig:emb}, where the projector was used directly from pretraining without alignment.

\model{} achieved an average \textit{k}-NN AUROC of $0.821$, compared to $0.851$ and $0.839$ for $2D$ PCA applied to ImageNet and SimCLR representations, respectively. While this reflects the expected cost of constraining the projection to $2D$ during pretraining, the learned representations nonetheless capture meaningful disease severity structure across datasets, without requiring any post-hoc dimensionality reduction. This cost is expected, since \textit{t}-SNE and PCA are applied post-hoc to frozen high-dimensional representations, where the model can encode multiple factors of variation without a 2D constraint. In contrast, \textit{t}-SimCNE's projector must compress these factors into 2D during the later stages of pretraining, while the encoder's representations are still shifting, making them a moving rather than fixed target. Additionally, rather than explicitly preserving the local geometry of the high-dimensional representations, \textit{t}-SimCNE's objective maximizes augmentation-invariant information. 

\begin{table}[b!]
\caption{\textit{k}-NN AUROC of $2D$ embeddings. For baselines, \textit{t}-SNE and PCA are applied post-hoc to high-dimensional representations. $^\dagger$~Aligned projector.}
\label{tab:glo_int}
\resizebox{\linewidth}{!}{%
    
    \begin{tabular}{@{}lcccccc@{}}
        \toprule
        \multicolumn{1}{c}{\textbf{Method}} &
        \textbf{ImageNet} &
        \textbf{SimCLR} &
        \textbf{\textit{t}-SimCNE} &
        \textbf{\textit{t}-SimCNE$^\dagger$} &
        \textbf{Dual-IFM} &
        \textbf{Dual-IFM$^\dagger$} \\ \midrule
        \textit{t}-SNE &
        0.904 &
        0.914 &
        \multirow{2}{*}{0.745} &
        \multirow{2}{*}{0.773} &
        \multirow{2}{*}{0.821} &
        \multirow{2}{*}{0.832} \\
        PCA $2D$ &
        0.851 &
        0.839 &
        &
        &
        &
        \\ \bottomrule
    \end{tabular}
}
\end{table}

\subsection{Local interpretability through class evidence maps}
\begin{figure}[t]
    \centering
    \includegraphics[width=0.8\textwidth]{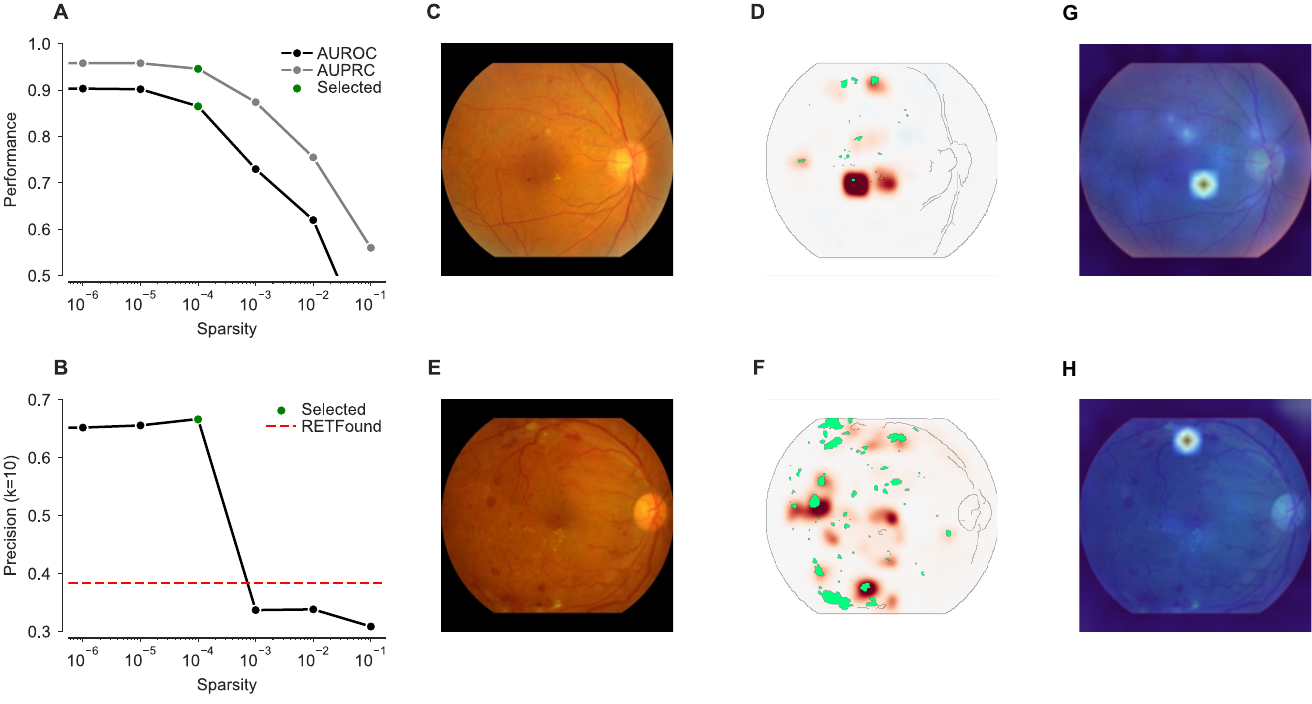}
    \caption{\textbf{Local interpretability}. (A) Fine-tuning performance and (B) localization precision vs. sparsity. (C, E) Example fundus images with (D, F) class evidence maps of \model{} and (G, H) LRP heatmaps from RETFound. Lesions overlaid in green.}
    \label{fig:localinterp}
\end{figure}

We evaluated the class evidence maps provided by \model{} for individual CFPs through the BagNet architecture~(Fig.~\ref{fig:localinterp}). We incorporated a sparsity constraint in the fine-tuning loss, as proposed by~\cite{donteu2023sparse}, which improves the localization of the evidence maps but affects the classification performance at higher values. We selected the sparsity value that maximized localization before performance deteriorated~(Fig.~\ref{fig:localinterp} A). To assess the localization precision of the evidence maps, we measured the overlap of the $k=10$ most activated patches with DR-related lesions on the IDRiD dataset, using \model{} fine-tuned for binary DR classification (healthy vs.\ diseased). \model{} achieved a precision of $0.674$, indicating that the sparsity constraint produced well-localized evidence maps, compared to $0.384$ for RETFound using post-hoc LRP~\cite{Chefer_2021_CVPR}. Qualitatively, the class evidence maps showed high activations in CFP regions with disease-related lesions~(Fig.\,\ref{fig:localinterp} C-F), with substantially higher overlap than LRP heatmaps from RETFound~(Fig.\,\ref{fig:localinterp} G,H). Together, these results demonstrate the ability of \model{} to provide faithful and clinically meaningful local explanations without the need for post-hoc attribution methods.

% \newpage
\section{Discussion}
We presented \model, an inherently interpretable foundation model for CFPs, that offers local interpretability and a direct visualization of the representation space for dataset-level exploration of learned structure. It achieved competitive performance across CFP benchmarks, demonstrating high-quality representations (linear probing) and adaptability to downstream tasks (fine-tuning). In some tasks, \model{} with 18.3M parameters surpassed RETFound, even though RETFound is a much larger model with over 303M parameters trained on 1.6M retinal images ($2\times$ our training data), resulting in $1.7\times$ the inference time (RETFound: 140 ms vs. \model: 81 ms). However, due to the BagNet architecture's small receptive field and the resulting large spatial feature maps, \model{} has a high memory footprint relative to parameter count (1.12 GB vs. RETFound's 1.31 GB), as detailed in Table~\ref{tab:compute-cost}. 

Crucially, \model's BagNet-based architecture provides inherent interpretability through class evidence maps that highlight lesion-relevant regions, unlike models like RETFound which rely on post-hoc attribution methods. These properties make \model{} well-suited for clinical deployment, where both accuracy and transparency are required. Additionally, training a BagNet is memory-intensive, requiring a relatively small batch size despite its low parameter count (Table ~\ref{tab:compute-cost}). Therefore, the availability of pretrained weights such as \model{} that improve downstream performance is advantageous.

Our work was limited to CFP and a restricted hyperparameter search during fine-tuning. Future work could extend to additional modalities, interpretable-by-design backbones and SSL objectives that could improve performance from pretraining. The inherent interpretability would make \model{} uniquely suited for a multimodal setting, while enabling dataset-level exploration across modalities. In summary, our results suggest that local interpretability and visualizations of the representation space can be obtained simultaneously without compromising representation quality in foundation models for medical images.

% \section*{Code availability}
% The code used for the study is available at \href{www.github.com}{www.github.com/}
\begin{credits}
    \subsubsection{\ackname} This project was supported by the Hertie Foundation and by the Deutsche Forschungsgemeinschaft under Germany’s Excellence Strategy with the Excellence Cluster 2064 “Machine Learning – New Perspectives for Science”, project number 390727645. PB is a member of the Else Kröner Medical Scientist Kolleg “ClinbrAIn: Artificial Intelligence for Clinical Brain Research”. The authors thank the AREDS Research Group for their contribution to this research with funding support provided by the National Eye Institute (N01-EY-0-2127). This research has been conducted using the UK Biobank Resource under Application Number (121699).
    \subsubsection{\discintname} The authors declare no competing interests.
\end{credits}

\begin{comment}  %% removed for anonymized MICCAI 2025 submission.
    
    % The following acknowledgement and disclaimer sections should be removed for the double-blind review process.  
    % If and when your paper is accepted, reinsert the acknowledgement and the disclaimer clause in your final camera-ready version.

\begin{credits}
\subsubsection{\ackname} A bold run-in heading in small font size at the end of the paper is
used for general acknowledgments, for example: This study was funded
by X (grant number Y).

\subsubsection{\discintname}
It is now necessary to declare any competing interests or to specifically
state that the authors have no competing interests. Please place the
statement with a bold run-in heading in small font size beneath the
(optional) acknowledgments\footnote{If EquinOCS, our proceedings submission
system, is used, then the disclaimer can be provided directly in the system.},
for example: The authors have no competing interests to declare that are
relevant to the content of this article. Or: Author A has received research
grants from Company W. Author B has received a speaker honorarium from
Company X and owns stock in Company Y. Author C is a member of committee Z.
\end{credits}

\end{comment}

% END OF 8th PAGE.

% 2 pages of references
\bibliographystyle{splncs04}
\bibliography{references.bib}

\end{document}